\documentclass{article}

\usepackage{arxiv}

\usepackage[utf8]{inputenc} % allow utf-8 input
\usepackage[T1]{fontenc}    % use 8-bit T1 fonts
\usepackage{hyperref}       % hyperlinks
\usepackage{url}            % simple URL typesetting
\usepackage{booktabs}       % professional-quality tables
\usepackage{amsfonts}       % blackboard math symbols
\usepackage{nicefrac}       % compact symbols for 1/2, etc.
\usepackage{microtype}      % microtypography
\usepackage{lipsum}		% Can be removed after putting your text content
\usepackage{graphicx}
\usepackage{natbib}
\usepackage{doi}

\title{Exploring large language models to generate Easy to Read content}

\author{ \href{https://orcid.org/0000-0003-3013-3771}{\includegraphics[scale=0.06]{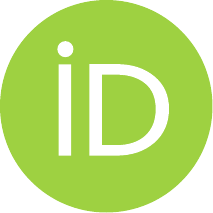}\hspace{1mm}Paloma Martínez}\\
	Computer Science and Engineering Department\\
	Universidad Carlos III de Madrid\\
	   Leganés, Madrid, Spain \\
	\texttt{pmf@inf.uc3m.es} \\
	\And
	\href{https://orcid.org/0000-0002-9021-2546}{\includegraphics[scale=0.06]{orcid.pdf}\hspace{1mm}Lourdes Moreno} \\
	Computer Science and Engineering Department\\
	Universidad Carlos III de Madrid\\
	 Leganés, Madrid, Spain \\
	\texttt{lmoreno@inf.uc3m.es} \\
	\And
	{\hspace{1mm}Alberto Ramos} \\
	Computer Science and Engineering Department\\
	Universidad Carlos III de Madrid\\
	 Leganés, Madrid, Spain \\
	\texttt{albramos@pa.uc3m.es}
}

\begin{document}
\maketitle

\begin{abstract}
Ensuring text accessibility and understandability are essential goals, particularly for individuals with cognitive impairments and intellectual disabilities, who encounter challenges in accessing information across various mediums such as web pages, newspapers, administrative tasks, or health documents. Initiatives like Easy to Read and Plain Language guidelines aim to simplify complex texts; however, standardizing these guidelines remains challenging and often involves manual processes. This work presents an exploratory investigation into leveraging Artificial Intelligence (AI) and Natural Language Processing (NLP) approaches to systematically simplify Spanish texts into Easy to Read formats, with a focus on utilizing Large Language Models (LLMs) for simplifying texts, especially in generating Easy to Read content. The study contributes a parallel corpus of Spanish adapted for Easy To Read format, which serves as a valuable resource for training and testing text simplification systems. Additionally, several text simplification experiments using LLMs and the collected corpus are conducted, involving fine-tuning and testing a Llama2 model to generate Easy to Read content. A qualitative evaluation, guided by an expert in text adaptation for Easy to Read content, is carried out to assess the automatically simplified texts. This research contributes to advancing text accessibility for individuals with cognitive impairments, highlighting promising strategies for leveraging LLMs while responsibly managing energy usage.
\end{abstract}

% keywords can be removed

\section{Introduction}

In a society increasingly saturated with information, the ability to understand digital content has become a real challenge for many people. Despite the widespread access to information facilitated by Information and Communication Technologies (ICT), a considerable number of people encounter difficulties in understanding textual content because not everyone can read fluently, and the information written can exclude many people.

The challenge of understanding texts containing long sentences, unusual words, and complex linguistic structures can pose significant accessibility barriers. The groups of users directly affected include people with intellectual disabilities and individuals with cognitive impairments, but it also impacts those with literacy or comprehension problems, the elderly, the illiterate, and immigrants whose native language is different.

In Spain alone, more than 285,684 individuals with intellectual disabilities face challenges in understanding texts not tailored to their needs \cite{Imserso}. Additionally, over 435,400 individuals with acquired brain injury (ABI) encounter obstacles in reading comprehension, often as a result of strokes or traumatic brain injuries \cite{INE2022}. The aging population adds another layer to this issue. With nearly 20\% of the Spanish population being over 65 years old and a global trend toward an aging population \cite{WHOAgeingHealth}, the need to adapt textual content to the cognitive needs of older individuals is clear. These individuals may experience a natural decline in reading comprehension abilities. By 2050, the global population over 60 years is expected to almost double, emphasizing the urgency to address cognitive accessibility barriers. Also, according to \cite{WHO2019DementiaGuidelines}, with approximately 50 million people affected by dementia worldwide, and a new case every three seconds, the number of people with dementia is expected to triple by 2050. Furthermore, individuals with low educational levels face more significant challenges. Despite over 86\% of the world's population being able to read and write, disparities in reading comprehension remain profound. This is evident in the PIAAC results, which reveal that the Spanish population aged 16 to 65 scores below the OECD and EU average in reading comprehension. These statistics underscore the significant impact of cognitive accessibility barriers on various reader groups and the need to provide simplified adapted texts with relevant information to citizens.

To provide universal access to information and make texts more accessible, there are initiatives like the Easy to Read and Plain Language guidelines. However, standardizing these guidelines is challenging due to the absence of tools designed to systematically support the simplification processes. Websites that offer simpler versions of texts currently rely on manual processes. As a solution, there are methods of Artificial Intelligence (AI) and Natural Language Processing (NLP) using language resources and models. 

This work presents an exploratory study on how approaches from these disciplines can be utilized to support the systematic fulfillment of simplifying the complexity of Spanish texts, with key premises being the use of easy reading guidelines and resources, and ensuring the participation of individuals with disabilities or experts in the field. in light of the high energy consumption of large generative AI models, which is not environmentally sustainable, and their associated costs that could increase social inequalities, this work aims to explore more cost-effective and energy-efficient solutions.

 The article is organized as follows: Sections 1 and 2 explain the motivation behind this research, as well as state-of-the-art NLP techniques applied to text simplification, with a special focus on creating Easy to Read content. Sections 3 and 4 cover the proposed methodology, including the datasets used in training and testing, the technical architecture deployed to generate Easy to Read content, the LLMs explored, and the experimental configurations. Sections 5 and 6 provide a preliminary analysis, as well as some insights.

\section{Background}
This section encompasses a discussion on strategies aimed at making text understanding and accessible, detailing their specific implementation in the Spanish language and alignment to standards within Spain, alongside a review of research in text simplification, especially within the realms of NLP and AI.

\subsection{Understanding Strategies: Easy to Read and Plain Language}
Historically, Inclusion Europe crafted standards for those with intellectual disabilities, emphasizing simplicity and the importance of feedback to ensure clarity, a cornerstone of the Easy to Read method for making information understandable to this group \cite{Freyhoff1998MakeItSimple}. In contrast, the International Federation of Library Associations and Institutions (IFLA\footnote{https://www.ifla.org/}) promoted broader, more adaptable guidelines, reflecting an understanding of accessibility for diverse reading levels. This approach broadened the scope of information accessibility. As time passed, the Easy to Read methodology evolved \cite{Nomura2010}, continually refining its strategies to serve the specific needs of individuals with intellectual disabilities.

Another approach is Plain Language, which historically predates Easy to Read and has been used in some fields, such as legal scope \cite{Wydick1979}, but has experienced a resurgence in recent times. The revival of Plain Language underscores its growing significance in accessible communication, reflecting an expanded recognition of the need for clear, understandable information \cite{Nomura2010}, \cite{EuropeanCommission2011}. Emphasizing direct, concise communication, Plain Language aims to be accessible to a wide audience, including those with cognitive impairments and the general public, like individuals with limited reading skills or non-native speakers. The goal of Plain Language is driven by an increased understanding of its role in promoting transparency. Various sectors such as government, healthcare, legal, and businesses are adopting Plain Language to improve communication with their stakeholders, highlighting its universal importance in creating an inclusive society where information is accessible to everyone, irrespective of their reading level or background.

The comparison between Easy to Read and Plain Language can be understood through several key aspects, each catering to different needs and audiences. Easy to Read is primarily designed for individuals with cognitive or intellectual disabilities, including those facing learning difficulties. This approach places a high emphasis on structural and linguistic simplicity, aiming to reduce cognitive load by using short and simple sentences, clear and direct language, and incorporating specific images and symbols to aid comprehension. A distinctive feature of Easy to Read is its practice of involving the target audience in testing and reviewing texts to ensure clarity and ease of understanding, directly obtaining feedback to refine the content. In contrast, Plain Language is directed toward a broader audience, including the general population, people with reading limitations, and non-native speakers, making it applicable across diverse groups. Its focus is on clarity, conciseness, and logical organization, employing techniques such as reader-centered organization, active voice, and the use of common and everyday words to eliminate ambiguity. This approach aims to make information accessible to as many readers as possible, with less emphasis on visual aids and a strong focus on avoiding jargon to promote general accessibility. 

The advantages of each approach are notable. Easy to Read's tailored support for individuals with specific needs ensures that content is accessible to those who face significant challenges in reading and understanding standard texts. The use of visuals further enhances comprehension, making it a highly effective method for its target audience. On the other hand, Plain Language's broad applicability ensures that a wide range of readers, regardless of their reading abilities or linguistic backgrounds, can access and understand information easily. Its principles of clarity enhance understanding for all, making it a tool for promoting clear communication.

The application domains also differ, with Easy to Read commonly used in educational materials, and legal and government documents designed for people with intellectual disabilities, while Plain Language is widely used in government communications, legal, health, educational documents, business documents, and websites aimed at a general audience.

Both approaches strive to make information accessible to all, highlighting the need for clear communication to foster a more inclusive society. However, it's important to note that simply applying some of their guidelines does not automatically qualify a text as Easy to Read or Plain Language. A text must adhere to all their respective guidelines to be truly considered as such. 

\subsection{Easy to Read in Spain}
In Spain, the significant advancement in accessible communication was highlighted by the approval and publication of the world's first Easy Read standard, \cite{UNE153101_2018}. Based on this standard, the Easy to Read methodology is built upon three key principles: treating Easy to Read as a method for publishing documents that address both writing and design, involving the target group in the creation process to ensure the final product meets their needs, and defining Easy to Read as a tool for people with reading difficulties rather than for the general populace.

A literature review on text simplification in Spain shows limited research aimed specifically at producing Easy to Read texts following these principles. Relevant initiatives that utilize AI and NLP methods, such as the Simplext project \cite{saggion2015making}, which aimed to develop an automated Easy Language translator, and the Flexible Interactive Reading Support Tool (FIRST) \cite{barbu2015language}, focused on text simplification for improved accessibility, demonstrate some progress in this area. However, despite the exploration of Easy to Read best practices, the full implementation of writing guidelines for complete adaptation to Spanish standards is rare. Additionally, much of the research on text simplification tends to overlook the participation of individuals with disabilities \cite{alarcon2023easier}. These efforts mainly concentrate on writing guidelines—spelling, grammar, vocabulary, and style—and less on document design and layout. It is worth mentioning previous work by the authors on a lexical simplification tool that did address enhancing simplification with visual elements like pictograms and provided simple definitions and synonyms in a glossary format (EASIER) \cite{alarcon2021lexical} \cite{Moreno2023} \cite{Alarcon2022}. Other works focus on supporting Easy to Read experts in their daily task of text adaptation \cite{SuarezFigueroa2022}.

This research aims to bridge these gaps by following the principles outlined in the UNE15310: 20181 standard, explicitly focusing on rigorously involving people with intellectual disabilities and individuals with cognitive impairments, in addition to addressing the writing-related guidelines. Our goal is to provide comprehensive support for the practical application of Easy to Read in Spain, covering both the simplification processes and the professional needs of Easy to Read experts.

\subsection{Related Work}
In the last years, the community of NLP and AI researchers addressed solutions to automatically translate texts into simpler ones but there are no approaches to automatically generate Easy to Read content from texts. Generation of this type of content is a complex task because the requirements of the target audience (people with intellectual disabilities) should be taken into account should be taken into account as introduced. 

Simplification can be approached as an all-in-one process or as a modular approach. In a modular approach, there is a pipeline composed of separated processes, for instance, replacing complex words with simpler ones, dividing coordinated or relative sentences into simple sentences, replacing passive sentences with active ones, etc. In an all-in-one approach simplification is done in a single step, such as those based on generative models. First previous works were focused on rule-based simplification approaches to reduce syntactical complexity in subordinate or coordinated sentences once a morpho-syntactic analysis is done using a set of hand-crafted rules, such as \cite{siddharthan2006syntactic}. Additionally, other approaches used machine learning approaches to discover patterns in parallel corpora following statistical machine translation such as \cite{specia2010translating} and \cite{coster2011simple}. See for a detailed survey on text simplification following these approaches. More recently, neural text simplification models such as transformed-based approaches have emerged. Research focus on encoder-based models, like BERT, that is pretrained to predict a word given left and right context using a high volume of text data. Other additional training methods can be done, for instance, next sentence prediction. These pretrained tasks can be leveraged to simplify words in a text or simplify complex sentences, \cite{qiang2020lexical}, \cite{martin2019controllable}.  

Traditional machine learning approaches or fine-tuning of Large Language Models (LLM) require datasets of labeled texts, and easy dictionaries among others used in training and testing systems. Manual annotation is costly and time-consuming.  In addressing the need for comprehensive resources, we conducted a survey of corpus resources in the Spanish language. This effort highlights different initiatives in the field of lexical simplification in Spanish. Parallel corpora, which include both original texts and their simplified versions, are extremely valuable tools for training text simplification algorithms, especially in languages with limited resources, such as Spanish. The most common corpora are those composed of a set of original sentences and their aligned simplified versions. As a result of this literature review, Table \ref{tab: resources corpora} compiles some corpora and datasets for text simplification in Spanish describing the source, type, annotators, and size.

Focusing on lexical simplification, three works on the Spanish language are highlighted: (1) in \cite{bott2012can}, an unsupervised system for lexical simplification in Spanish, which utilizes an online dictionary and the web as a corpus is introduced.  Three features (word vector model, word frequency, and word length) are calculated to identify the most suitable candidates for the substitution of complex words. The combination of a set of rules and dictionary lookup allows for finding an optimal substitute term. (2) \cite{ferres2017spanish} describes the TUNER Candidate Ranking System, an adaptation of the TUNER Lexical Simplification architecture designed to work with Spanish, Portuguese, and English. This system simplifies words in context, omitting the complete simplification of sentences. The system proposes four phases:  sentence analysis, disambiguation of word senses (WSD), synonym classification based on word form frequency using counts from Wikipedia in the respective language, and morphological generation. (3)  Our previous work in \cite{alarcon2021lexical}, a neural network-based system for lexical simplification in Spanish that uses pre-trained word embedding vectors and BERT models is described. These systems were evaluated in three tasks: complex word identification (CWI), Substitute Generation (SG), and Substitute Selection (SS). In the case of the CWI task, the shared tasks dataset for CWI 2018 \cite{yimam2017multilingual} in Spanish was used. For SG and SS tasks, the evaluation was carried out using the EASIER-500 corpus \cite{alarcon2023easier}. The fourth task, substitute classification (SR), was not evaluated due to the absence of Spanish datasets for lexical simplification that could be used for that purpose.

\begin{table*}[h]
\caption{Survey Study of Corpora for Text Simplification in Spanish}
    \centering
    \begin{tabular}{p{0.15\linewidth}|p{0.17\linewidth}|p{0.22\linewidth}|p{0.28\linewidth}}

     \textbf{Corpus} & \textbf{Type and source} & \textbf{ Annotators} & \textbf{Size} \\
     \hline
        Simplext, \cite{vstajner2015automatic}&News stories from Servimedia& Trained human editor &200 short news articles \\
        \hline
        RANLP 2017, \cite{yimam2017multilingual} & Wiki news and Wikipedia articles  & 54 turkers (Native and non-native speakers)&  14,280 sentences with target complex words \\
        \hline
        PPDB-S/M \cite{vstajner2019improving}& Texts from Europarl, Wikipedia and simple Wikipedia & Built by filtering and ordering paraphrases pairs from PPDB &  5,709 unigrams for Small (S) DB size and 15,524 unigrams for M (Larger DB) size \\
        \hline
        CASSA \cite{vstajner2019improving}&OpenThesaurus and EuroWordnet &  All unique 5-grams pairs from CASSA& 5,640,694 5-grams \\
        \hline
        CWI 2018 \cite{yimam2017multilingual}&  WikiNews and Wikipedia &  10 annotators, a mix of native and non-native speakers & 17605 annotated words \\
        \hline
        ALexS 2020 \cite{ortiz2020overview} &  Transcriptions of academic videos &  430 students&  55 texts with 9,175 words, 723 annotated as complex words\\
        \hline
        EASIER \cite{alarcon2023easier}  &  News articles &  3 native Spanish annotators (people with intellectual disabilities, expert linguists and specialists in Easy to Read)& 3977 sentences annotated with 8155 complex words and 3396 sentences annotated with 7892 suggested synonyms (121,064 words)\\
        \hline
        Newsela \cite{xu2015problems}&  News articles &  Manually produced by professional editors& Parallel simple - complex articles with 11-grade levels (1,130 articles) \\
        \hline
        FinTOC 2022 \cite{el2022financial}&  Financial documents (FinT-esp) & Manual processing by annotators & 90 financial documents with an average of 148 tags per document (250,000 words)\\
        \hline
        TSAR \cite{vstajner2022lexical}& WikiNews and Wikipedia (CWI 2018) &   Prolific\footnote{https://www.prolific.com/} annotators&  368 instances with the sentences target complex words, and gold annotations (31,021 words)\\
        \hline
        ALEXSIS \cite{ferres2022alexsis}& CWI Shared Task 2018 Spanish dataset (News texts) &  Prolific  annotators& 381 sentences with a target complex word, and 25 candidate substitutions \\
        \hline
        CLARA-MeD \cite{campillos2022clara}&  Drug leaflets, summaries of products, abstracts, cancer-related summaries, and clinical trials& Manually annotated by pairs of expert annotators& A collection of 24,298 pairs of professional and simplified texts (96M tokens)\\
    \end{tabular}
    \label{tab: resources corpora}
\end{table*} 

After analyzing Easy to Read guidelines and as a result of this analysis, we present in Table \ref{tab: Guidelines Easy to Read} a proposed set of Easy to Read guidelines based on the Spanish standard and European Guidelines in Spanish \cite{ilsmh1998camino}. This proposed set of guidelines drives the automated adaptation process, influencing the design of prompt inputs for the models and also serving as a benchmark for the validation of the experimentation. This work aims to fine-tune decoder-based models such as Llama2 \cite{touvron2023llama} using datasets composed of Easy to Read Spanish documents as a proof of concept in generating Easy to Read content.

\section{Method}

In this section, the methodology of this research is described, including the dataset used to train and test the generative LLMs to simplify texts in Easy to Read formats, the method used to align the original sentences with the corresponding simplification (if exists), the LLM architecture chosen and finally the evaluation procedure. The use of an open-source generative model is proposed, highlighting its capacity to address complex language-related tasks. This approach involves fine-tuning the model using a corpus containing texts adapted for easy reading, to achieve specific adaptation to readability requirements, primarily targeting individuals with intellectual disabilities and difficulties in reading comprehension. 
The contributions of this work are highlighted in this section, including the creation of a parallel corpus resulting from an alignment process, which serves as a resource of immense value for future research in text simplification in Easy to Read formats. The use of an open-source generative model is emphasized, which is adapted through a fine-tuning process to generate Easy to Read content.

\subsection{Dataset}
The Amas Fácil Service\footnote{https://amasfacil.org/}, an entity that offers a specific service of text adaptation for easy reading in addition to other cognitive accessibility services for individuals with intellectual disabilities, has provided various sets of general domain documents. These documents feature the original text and its corresponding adaptation for easy reading, tailored for users with intellectual disabilities who encounter challenges in reading comprehension. This dataset emerges as a significant source of data in the simplification. These texts are adapted for users with intellectual disabilities who experience reading comprehension difficulties. From these texts, a subset of 13 documents related to sports guides, literature, competitive examinations, and exhibitions has been selected.

The corpus has a total of 1941 sentences, 56,212 words of the original text and 40,372 words of adapted texts, with an average of 27 words in the source sentence and 20 words in an adapted sentence. The original texts contain terms and expressions that impede comprehension for individuals with intellectual disabilities.

The original source texts and their respective adaptations follow a different language and style depending on the topic of the text. For instance, the language used in literature texts is entirely different from that employed in a sports guide. We have employed texts from different themes because there is a limitation regarding the number of texts adapted for easy reading in Spanish. Conversely, more resources are available for texts with lexical simplification and adapted to plain language, but this deviates from the focus on accessibility and inclusion for individuals with intellectual disabilities.

\subsection{Text selection}
In the methodology employed for sentence selection, initially, any sentences from the original texts that did not have an easy reading adaptation were excluded. This exclusion ensures that only texts with both the original and adapted versions are considered for analysis. In our approach, each original sentence is uniquely identified and paired with its potential adaptations. These pairs are then evaluated based on the cosine similarity between the original sentence and its adaptation. As part of the cosine similarity evaluation process, any original sentences lacking a corresponding candidate adaptation are removed.

Subsequently, from the pool of candidate adaptations for each original sentence, we select the adaptation with the highest cosine similarity score. This process guarantees that for each original sentence, we identify the most closely aligned adapted version for easy reading. This selection strategy results in a curated set of sentence pairs, each consisting of an original sentence and its optimal easy reading adaptation. Table \ref{tab:theme} provides an overview of the distribution of texts and sentences across various topics, illustrating the breadth of content considered in our analysis.

Adapting texts to enhance readability often involves significant simplification, including the omission of words, entire sentences, or parts thereof. This level of modification introduces challenges in accurately identifying corresponding sentences between the original and adapted versions. To overcome this challenge and facilitate the collection of parallel data, we have developed a specialized sentence aligner which is presented in the following section. 

  \begin{table}[h]
    \centering
    \caption{Distribution of texts by theme}
    \label{tab:theme}
    \begin{tabular}{ccc}
      \textbf{Theme} & \textbf{Total text} & \textbf{Total sentences}\\ \hline
      Sport           & 3    &   480 \\             
      Literature      & 2    &     1118 \\               
      Exhibitions     & 2   &   72   \\               
      Competitive examinations     & 5     &     271 \\ \hline
    \end{tabular}
  \end{table}

\subsection{Sentence Alignment Process}
The adaptation of texts to enhance readability often involves significant simplification, including the removal of words, entire sentences, or parts thereof. This level of modification introduces challenges in accurately identifying which sentences correspond between the original text and its adapted version. To address this issue and facilitate the collection of parallel data, we have developed a sentence aligner.

The main function of this aligner is to process sentences from the texts, aligning them between the original text and its adapted version for easy reading. Initially, the texts are segmented into individual sentences to generate sentence embeddings. These embeddings are vector representations that encapsulate semantic information about the sentences, enabling the comparison and measurement of semantic similarity between sentences. For generating these embeddings, we utilize the Sentence Transformers framework \cite{reimers-2019-sentence-bert}, which offers a suite of pre-trained models specifically optimized for the task of sentence alignment.

Upon generating the sentence embeddings, we determine the semantic similarity between them by computing the cosine similarity. The LaBSE model \cite{feng2020language}, a BERT \cite{devlin2018bert} modification, is employed for this purpose. Optimized to produce analogous representations for pairs of sentences, this model assesses the cosine similarity across all sentence pairs, selecting the most similar sentence among potential candidates as the correct alignment.

It's important to highlight that the development of this sentence aligner and the subsequent creation of a parallel corpus represent significant contributions to this research. The parallel corpus, which is in the process of being published, will offer a valuable resource to the scientific community, aiming to enhance text simplification efforts.

\subsection{Generating Easy to Read texts}
In line with our initial commitment to address the environmental and economic impacts of large generative AI models, this section demonstrates a practical application that aligns with our goals of sustainability and inclusivity. The reliance on energy-intensive models poses significant environmental challenges and exacerbates social inequalities through their substantial monetary costs. To mitigate these issues, we have selected Llama2, an open-source \cite{touvron2023llama} model provided by Hugging Face (Llama-2-7b-hf), for its efficiency and adaptability in generating Easy to Read content. Also, it has shown considerable promise in executing complex reasoning tasks across a broad spectrum of domains, from general to highly specialized fields, including text generation based on specific instructions and commands.

The auto-regressive Llama2 transformer is initially trained on an extensive corpus of self-generated data and then fine-tuned to human preferences using Reinforcement Learning with Human Feedback (RLHF) \cite{touvron2023llama}. While the training methodology is simple, the high computational requirements have constrained the development of LLMs. Llama2 is trained on 2 billion tokens of text data from various sources and has models ranging from 7B to 70B parameters. Additionally, they have increased the size of the pretraining corpus by 40\%, doubled the model context length to 4096 tokens, and adopted clustered query attention to enhance the scalability of inference in the larger 70B model.

Since Llama2 is open-source and has a commercial license, it can be used for a multitude of tasks such as lexical simplification. However, to achieve better results and performance in the task, it is necessary to fine-tune it with the specific data and adapt it to the readability needs of individuals with intellectual disabilities. Therefore, the weights and parameters of the pre-trained model will be adjusted, resulting in increased precision in task outcomes, as well as a reduction in the likelihood of inappropriate content.

A key contribution of this work is the demonstration of the adaptability of Llama2 for the generation of Easy to Read texts, marking a significant step towards customizing advanced language models for enhanced accessibility.

\subsection{Evaluation procedure}

In validating our model, we diverge from using traditional performance metrics common in text simplification tasks, such as BLEU \cite{papineni2002bleu}, SARI \cite{xu-etal-2016-optimizing}, and the Flesch-Kinkaid readability index \cite{kincaid1975derivation}, among others. Instead, we prioritize a qualitative human evaluation conducted by professional easy reading adapters and individuals with intellectual disabilities. While traditional metrics can gauge a model's effectiveness to some extent, they do not adequately capture the nuances of language understanding, simplicity, and the preservation of meaning in practical contexts.

Engaging professional easy reading adapters in our evaluation process is critical to ensuring the results are genuinely accessible and beneficial for people with intellectual disabilities. This human-centric approach is vital in the context of easy reading, where adherence to specific guidelines is necessary to enhance readability and understanding.

For the evaluation, we have chosen documents related to indoor soccer regulations and sports guidelines, specifically focusing on the competition regulations for indoor soccer and the sports guide related to the Organic Law for Comprehensive Protection of Children and Adolescents against Violence (LOPIVI) within the sports domain. These documents provide a suitable basis for assessing if the model's outputs align with Easy to Read standards.

The qualitative evaluation conducted by experts is based on their expertise in adapting texts in easy reading and the standards outlined in Table \ref{tab: Guidelines Easy to Read}, which summarizes the guidelines for adapting texts Easy to Read as introduced in \cite{ilsmh1998camino} and \cite{de2018norma}. This table is the result of a comprehensive analysis of easy reading guidelines, drawn from standards and best practices.

\begin{table*}[h]
\caption{ Easy to Read guidelines}
    \centering
    \begin{tabular}{p{0.05\linewidth}|p{0.6\linewidth}|p{0.1\linewidth}|p{0.1\linewidth}}

     \textbf{ID} & \textbf{Guideline} & \textbf{ID UNE} & \textbf{ID UE} \\
     \hline
      
        G1 & One should not write words or phrases with all their letters in uppercase, except when they are acronyms  & 6.1.1 &  - \\
        \hline
        G2 & Linked ideas should be separated by a period instead of a comma  & 6.1.4 & 5.13 \\
        \hline
        G3& The semicolon (;) should not be used  & 6.1.7 & 5.13 \\
        \hline
        G4 & Use simple and commonly used language  & 6.2.1 & 5.1 \\
        \hline
        G5 &  Vocabulary should be appropriate for the end user of the document. & 6.2.2 & 5.4 \\
        \hline
         G6&  Avoid using abstract, technical, or complex terms &  6.2.4 & 5.2 \\
        \hline
        G7& Avoid superlatives  & 6.2.8 & - \\
        \hline
        G8&  Avoid using words in other languages unless they are widely known and properly explained & 6.2.10 & 5.17 \\
        \hline
       G9 &  Avoid abbreviations & 6.2.11 & 5.20 \\
          \hline
         G10& Deter from using expressions or metaphors that all readers may not understand unless they are common in everyday language  & 6.2.15 & 5.15  \\
        \hline
        G11 &  Use the same word throughout the text to refer to the same object or referent & 6.2.17& 5.12 \\
        \hline
        G12 & Use simple sentences and avoid complex sentences  & 6.3.1 & 5.7 \\

         \hline
        G13 &  Use the present indicative whenever possible & 6.3.2 & -\\
         \hline
        G14 &  One should avoid compound or uncommon verb tenses, as well as the use of conditionals and subjunctives & 6.3.3& 5.14 \\
        \hline
        G15 &  Avoid using the passive voice & 6.3.4 & 5.10 \\
        \hline
        G16 &  Use the imperative only in clear contexts to avoid confusion with the third person singular of the present indicative & 6.3.6 &  -\\
        \hline
            G17&One should avoid the use of impersonal sentences&6.3.7& 5.6\\
        \hline
       G18 &  One should avoid using two or more verbs in a row, except for periphrases with modal verbs like "should," "want," "know," and "can." & 6.3.9& 5.1  \\
        \hline
        G19& Preferably use affirmative sentences, except in cases such as simple prohibitions, where negative forms may be clearer and more direct & 6.3.10 & 5.9 \\
        \hline
       G20  &  Avoid negative forms and double negations & 6.3.11& 5.9 \\
        \hline
         G21 & Include only one main idea in each sentence & 6.3.15& 5.8 \\

    \end{tabular}
    \label{tab: Guidelines Easy to Read}
\end{table*}

\section{Experimental setup}

\begin{figure}
    \centering
       \includegraphics[width=1\textwidth]{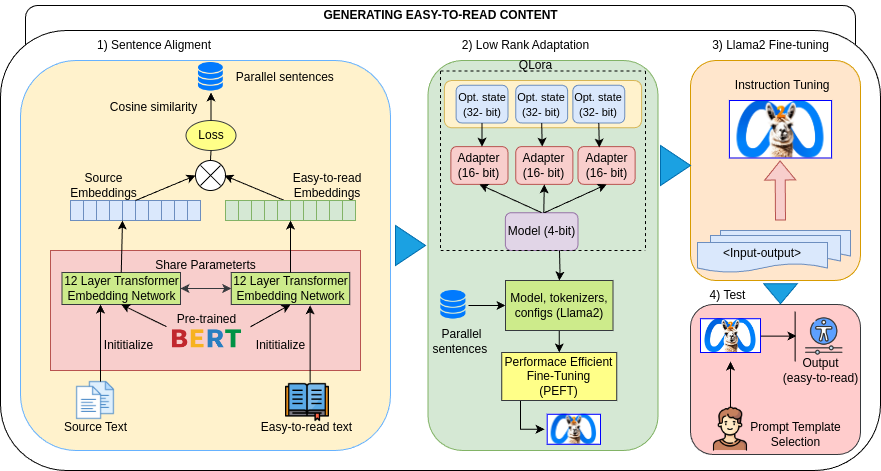} 
    \caption{Diagram of the system architecture}
    \label{fig:Diagram}
\end{figure}

Building on the methodology introduced in the previous section, an architecture is depicted in Figure \ref{fig:Diagram}. This architecture outlines four key processes: (1) sentence alignment, (2) low-rank adaptation, (3) LLM fine-tuning for the smaller 7B parameter Llama2 model, and (4) testing utilizing both Llama2 7B and Llama2 70B models.

Sentence alignment is the process required to match sentences from source texts to the corresponding sentences of Easy to Read adaptation preserving the meaning across the texts used.  Sentences without correspondence are deleted, and among those with multiple candidates, the one with the highest cosine similarity to the original sentence is chosen. To achieve this goal, we employ the Language-agnostic BERT Sentence Embedding (LaBSE) encoder model \cite{feng2020language}, pre-trained to support 109 languages. The model has been trained with both Masked Language Modeling (MLM) and Translation Language Modeling (TLM), allowing it not to depend on parallel datasets for training heavily. It has an architecture similar to BERT \cite{devlin2018bert}. The model transforms sentences into vector representations, which are used to calculate the similarity between sentences.

Once the aligned corpus is obtained, the LLama2 model is used to perform the task of text simplification into Easy to Read text. There are some available Spanish LLMs at the Barcelona Supercomputing Center \cite{bsc-cns} but the available Spanish decoder-based GPT-2  \cite{gutierrezfandino2021spanish} is older than the new LLama 2 |\cite{touvron2023llama} that has emerged showing better results across a variety of tasks.

To fine-tune the Llama2 model with 70B parameters, significant computing resources are required. We have utilized the smaller Llama2 model with 7B parameters. Yet, it still requires 30 GB of GPU memory, so we employed the QLoRa (Quantitative Low-Rank Adaptation) \cite{dettmers2023qlora} technique to reduce the model size and expedite the process, achieving greater efficiency. QLoRA technique is employed using the PEFT library\cite{peft} to fine-tune LLMs with higher memory requirements combining quantization and LoRA, where quantization is the process of reducing the precision of 32-bit floating-point numbers to 4 bits. Precision reduction is applied to the model's parameters, layer weights, and layer activation values. This results in agility in calculations and improvement in the model's storage size in GPU memory. While achieving greater efficiency, there is a slight loss of precision in the model. The LoRa technique includes a parameter matrix in the model that helps it learn specific information more efficiently, leading to faster convergence.

The Llama2 7B model is fine tuned with the entire corpus dataset, a total of 1941 sentences. We implement it in PyTorch using the Transformers and QLoRa libraries. The entire implementation is completed using Jupyter Notebook, and an NVIDIA GeForce RTX 3060 12GB is used to train the text simplification model. The GPU facilitates asynchronous data loading and multiprocessing.

The Llama2 model is pre-trained with 32 layers and 7B parameters. The pre-trained model shows significant improvements over the previous version Llama1. It has been trained with 40\% more tokens and a context length of 4096 tokens. We train it for four epochs with a maximum length of 512 tokens for input to the transformer model. Table \ref{tab:hyperparameters} shows the hyperparameters of the model. After fine-tuning, a set of instruction templates is used to guide the system on the desired output. The prompt used is: \textit{Transform the sentence to make it easier to understand for people with intellectual disabilities and difficulties in reading comprehension. Use very simple, short, direct sentences in the active voice, and avoid complicated words}.

\begin{table*}[h]
\caption{ Hyperparameters of the LLM tested }
  \centering
  \begin{tabular}{cc}
    \textbf{Hyperparameter} & \textbf{Llama2}  \\\hline
    bnb 4bit compute dtype & float16 \\
    
    bnb 4bit quant type & nf4 \\
    
   cache & False \\
   
    lora alpha & 16 \\
   
    lora dropout & 0.1  \\
    
    lora r & 64 \\
    
    batch size & 6 \\
    
    optim & paged adamw 32bit  \\

     learning rate &  $2 \times 10^{-5}$  \\
    
    max grad norm & 0.3 \\
    
    warmup ratio & 0.03  \\

    max seq length & 512  \\ \hline
  \end{tabular}
  
  \label{tab:hyperparameters}
\end{table*}

To complement our main findings, additional experiments were conducted using various approaches to text simplification. Specifically, the \textit{ollama} library was employed to test the Llama2 model with 70B parameters without fine-tuning, contrasting it with our primary method that involves fine-tuning a smaller Llama2 model. This approach facilitates local model testing, leveraging 4-bit quantization and packaging model weights, configurations, and datasets into a unified Modelfile. Table \ref{tab:AlternativeAproaches} outlines these experiments, providing insights into the efficacy of different configurations and prompts in generating accessible texts. 

A particular focus of these experiments was the translation approach, where the original text is translated into English, simplified, and then the simplified text is translated back into Spanish. This method evaluates the impact of translations on the coherence and accessibility of the final simplified text. This aspect is crucial because it directly addresses the challenges posed by the limited presence of the Spanish language in the resources of existing LLMs. 

These comparative analyses aim to identify optimal strategies for simplifying texts into Easy to Read formats, taking into account available resources and the underrepresentation of the Spanish language in current LLMs resources.

\begin{table*}[h]
\caption{Experiments performed using Llama2 70B and Llama2 7B, different prompts and with/without prompt and output translations}
  \centering
  \begin{tabular}{p{0.03\linewidth}|p{0.08\linewidth}|p{0.07 \linewidth}|p{0.11\linewidth}|p{0.6\linewidth}}

    \textbf{ID} & \textbf{LLM}& \textbf{Fine-Tuning}& \textbf{Translation approach} & \textbf{Prompt}  \\\hline
    E1 & Llama2 7B&Yes&No& Short and straightforward prompt, indicating the main idea, the target user, and how the simplification should be (output: Use very simple, short, and direct sentences in active voice, avoiding complicated words) \\
    \hline
    E2 & Llama2 7B&Yes&Yes& Same E1 prompt, but translated into English. In this process, the model must first translate the input into English, then simplify the sentence, and finally translate the simplification into Spanish \\
    \hline
    E3 & Llama2 70B&No&No&Short and straightforward prompt, indicating the main idea, the target user, and how the simplification should be (output: Use very simple, short, and direct sentences in active voice, avoiding complicated words) \\
    \hline
    E4 & Llama2 70B&No&Yes&Same E3 prompt, but translated into English. In this process, the model must first translate the input into English, then simplify the input text, and finally translate the simplification into Spanish \\ 
    \hline
    E5 & Llama2 70B&No&No& The prompt begins by presenting the main idea and specifying the user to whom the simplification is directed. Finally, the guidelines that the model must follow to carry out the simplification are detailed, which are found in Table \ref{tab: Guidelines Easy to Read} \\

  \end{tabular}
  \label{tab:AlternativeAproaches}
\end{table*}

\section{Results and discussion}
This section presents an exploratory evaluation of the Easy to Read simplification system, results from specific experiments, an expert review of simplification quality, error classification, and a comparative analysis. It concludes with implications for future research.

\subsection{Overview}
We present an exploratory evaluation of our simplification system using documents on sports regulations and guidelines. Table \ref{tab:exampleSimplification} displays examples of sentences input into the system and their corresponding simplifications according to Easy to Read guidelines performed by an expert human adapter in Easy to Read.

\begin{table*}[h]
\caption{Examples of sentences used as input models and corresponding human-adapted versions}
    \centering
    \begin{tabular}{p{0.05\linewidth}|p{0.45\linewidth} |p{0.45\linewidth}}
     
     \textbf{ID} & \textbf{Input sentence} & \textbf{Human adaptation to Easy to Read guidelines} \\
     \hline 
        S1&\textit{Para la disputa de los encuentros, podrán convocarse un máximo de 14 jugadores. Dada la limitación del acta, los jugadores de más se añadirán en el reverso de esta y será reflejado por el árbitro}  (A maximum of 14 players may be called up for the matches. Due to the limitation of the score sheet, the extra players will be added on the back of it and will be recorded by the referee)& \textit{El equipo puede llamar a jugar hasta 14 jugadores para cada partido. En el acta en el que el árbitro inscribe a los jugadores solo caben 14. Cuando el equipo llama a jugar a más de 14 jugadores,el árbitro escribe sus nombres en la parte de atrás del acta} (The team can call up to 14 players to play each match. In the score sheet where the referee registers the players, only 14 can fit. When the team calls up more than 14 players to play, the referee writes their names on the back of the score sheet)\\ \hline
        S2&\textit{Los equipos podrán inscribir un mínimo de 8 deportistas y un máximo de 16 deportistas por equipo} (Teams may register a minimum of 8 athletes and a maximum of 16 athletes per team)& \textit{Los equipos pueden inscribir entre 8 y 16 deportistas por equipo} (Teams can register between 8 and 16 athletes per team)\\ \hline
        S3&\textit{Para la inscripción de equipos se realizará a través de la plataforma de Gestión de Licencias de Femaddi}  (Team registration will be done through the Femaddi License Management platform)& \textit{Los equipos y los deportistas deben inscribirse  en la plataforma de Gestión de Licencias de FEMADDI en la página web: https://femaddi.playoffinformatica.com/}(Teams and athletes must register on the FEMADDI License Management platform on the website: https://femaddi.playoffinformatica.com/)\\ 
     
    \end{tabular}
    \label{tab:exampleSimplification}
\end{table*} 

\begin{table*}[h]
\caption{Output of the experiments}
    \centering
    \begin{tabular}{p{0.11\linewidth}|p{0.08\linewidth}|p{0.7\linewidth}}
     \textbf{Experiment} & \textbf{Sentence} & \textbf{Output}  \\
     \hline
       E1  & S1&  \textit{ Los equipos pueden reunir 14 jugadores en cada partido. El árbitro añadirá  jugadores en la parte de atrás del acta cuando el equipo tenga más de 14 jugadores} (The teams can field 14 players in each match. The referee will add players to the back of the score sheet if the team has more than 14 players) \\
       \hline
       E2  & S1& \textit{Para los partidos, se pueden convocar un máximo de 14 jugadores.  Dado que hay 14 jugadores en el acta, los jugadores que sean más  se añadirán en el reverso del acta por el árbitro} (For the matches, a maximum of 14 players can be called up. Since there are 14 players on the score sheet, any additional players will be added to the back of the score sheet by the referee) \\
       \hline
       E3  & S1& \textit{Los equipos pueden llamar a 14 jugadores a jugar el partido. Si hay más jugadores, el arbitro los añadirá en la parte de atrás del acta} (The teams can call up 14 players to play the match. If there are more players, the referee will add them to the back of the score sheet) \\
       \hline
       E4  & S1& \textit{Se puede llamar a un máximo de 14 jugadores para el partido. Los jugadores adicionales se añadirán en la parte posterior de la lista y se informará al árbitro} (A maximum of 14 players can be called for the match. Additional players will be added to the back of the list, and the referee will be informed) \\
       \hline
       E5  & S1& \textit{En un partido pueden jugar hasta 14 jugadores. El árbitro añadirá  jugadores en la parte de atrás de la lista cuando el equipo tenga más de 14 jugadores} (Up to 14 players can play in a match. The referee adds players to the back of the roster when the team has more than 14 players.) \\
       \hline
       E1  & S2& \textit{El equipo puede inscribir entre 8 y 16 deportistas} (The team can register between 8 and 16 athletes) \\
       \hline
       E2  & S2& \textit{Los equipos pueden inscribir un mínimo de 8 deportistas y un máximo de 16 deportistas} (Teams can register a minimum of 8 athletes and a maximum of 16 athletes) \\
       \hline
       E3  & S2& \textit{Los equipos pueden inscribir entre 8 y 16 deportistas} (The teams can register between 8 and 16 athletes) \\
       \hline
       E4  & S2& \textit{Un equipo puede tener entre 8 y 16 deportistas} (A team can have between 8 and 16 athletes) \\
       \hline
       E5  & S2& \textit{Los equipos pueden inscribir entre 8 y 16 jugadores} (Teams can register between 8 and 16 players) \\
       \hline
       E1  & S3& \textit{ Los equipo deben inscribirse en la plataforma de Femaddi para poder participar} (The teams must register on the Femaddi platform in order to participate) \\
       \hline
       E2  & S3& \textit{Para inscribirse en la competición, hay que usar la plataforma de licencias de Femaddi} (To register for the competition, you must use the Femaddi licensing platform) \\
       \hline
       E3  & S3& \textit{Para inscribir equipos usaremos la plataforma de Femaddi} (To register teams, we will use the Femaddi platform) \\
       \hline
       E4  & S3& \textit{La plataforma de Gestión de  Licencias de Femaddi se utiliza para inscribir los equipos} (The Femaddi Licensing Management Platform is used to register teams) \\
       \hline
       E5  & S3& \textit{Los equipos deben inscribirse en la Plataforma de Licencias Femaddi} (Teams must register on the Femaddi Licensing Platform) \\

    \end{tabular}
    \label{tab:Output}
\end{table*}

The expert adapter conducts a thorough evaluation of the adaptation by both Llama2 models. In an initial review, a significant achievement in lexical simplification was evident. Afterward, Easy to Read guidelines have been analysed in detail to identify those that have been implemented and those that have not.

\subsection{Limitations of the Experimental Study}
In this experimentation, certain guidelines will not be evaluated due to their inapplicability to the texts under consideration. Specifically, guidelines G1, G3, G7, G9, and G16 cannot be assessed. G1 is irrelevant because the texts do not contain words or phrases entirely in uppercase letters, except for acronyms. G3 is not applicable as none of the input sentences include semicolons. G7, concerning the avoidance of superlatives, does not apply because the original texts do not use superlative forms. G9, which advises against abbreviations, is not relevant here as the texts do not contain any abbreviations. Lastly, G16, pertaining to the use of the imperative mood, is not applicable because the imperative is not used in any of the input sentences, making it impossible to assess the guideline's adherence in this context. Of the 21 guidelines provided, 16 will be applicable in the experimental study

\subsection{Experimental Results}
Table \ref{tab:Output} shows the result of the experiments detailed in Table \ref{tab:AlternativeAproaches} along with the corresponding input sentence provided in Table \ref{tab:exampleSimplification}.

The experiments explored different approaches to simplification, with a focus on assessing the effectiveness of text simplification for easy reading in specific knowledge areas where the model has been fine-tuned, the impact of translations on simplification quality, and the performance of simplification without domain-specific fine-tuning.

\subsection{Discussion}
The results of the experiments with different approaches are presented below, and detailed in Table \ref{tab:AlternativeAproaches}. 

\begin{itemize}
\item The objective of Experiment E1 is to confirm the effectiveness of text simplification for easy reading in specific knowledge areas where the model has been fine-tuned. In this case, the focus was on the sports domain, using sports regulations and guidelines as references. It was observed that the performance of simplification is notably better when the model is familiar with the domain.

\item In experiment E2, the default Llama2 7B parameter model achieves accurate translations from Spanish to English. However, after fine-tuning for easy reading simplification, the model does not accurately translate. In most cases, the outputs remain in Spanish, and some show simplifications, even though the original prompt only requests translation into English. This discrepancy may stem from the layer selection during the fine-tuning process, where weight modifications enable the model to adapt to the new task. In this specific process, all layers were modified since it is recommended to use the complete set of layers for more complex and demanding tasks in natural language processing. However, it is possible that using fine-tuning with 8 or fewer layers could achieve correct translation into English. So, the option of employing the method of translating into English, simplifying the text, and then translating it back into Spanish for the fine-tuned model is rejected. As shown in Table \ref{tab:Output}, this approach yields the most deficient results.

\item The results of experiment E3 are less satisfactory compared to model 7B with fine-tuning. This is because it is limited solely to replacing terms with simpler synonyms, without achieving shorter or more direct sentences. The reason is that the model lacks familiarity with examples of easy reading, has not been previously trained in this area, and does not possess the necessary guidelines to carry out proper simplification in easy reading.

\item In experiment E4, the results are superior to those obtained in experiment E3. The model effectively simplifies by using short and direct sentences, with a clear main idea and avoiding using complex words. Although not specified in the request, the model often separates sentences using periods instead of commas. The Llama2 7B model with fine-tuning yields better results in domains where it is familiar compared to the results obtained using this approach.

\item Finally, in experiment E5, the results can be considered superior to the 7B model with fine-tuning. This model follows the instructions and guidelines of Table \ref{tab: Guidelines Easy to Read} to perform Easy to Read simplification. Additionally, the 70B model with this prompt does not need to be familiarized with the domain like the 7B model with fine-tuning to achieve good results.
\end{itemize}

The table \ref{tab: ClassificationError} showcases the classification of errors identified during the evaluation of results by the Easy to Read adapter. It details the types of errors along with their corresponding descriptions and their alignment with the Easy to Read guidelines. Only errors found in the Llama2 7B model with fine-tuning and the Llama2 70B model using the prompt that includes easy reading guidelines are presented. These approaches have been demonstrated to yield the most competitive results.

\begin{table*}[h]
\caption{Classification of errors detected by the human evaluator}
    \centering
    \begin{tabular}{p{0.3\linewidth}|p{0.55\linewidth}}
     \textbf{Type} & \textbf{Error}  \\
     \hline
       Number and gender agreement & It's uncommon, but sometimes there are no gender and number agreements because multilingual LLMs are trained with few texts in Spanish compared to English\\
       \hline
        Use the same term & In some cases, the same word is not used throughout the text to refer to the same object or referent. It does not comply with guideline G11. \\
       \hline
         Explanation of terms & Some technical terms are not explained. It does not comply with guideline G6.\

    \end{tabular}
    \label{tab: ClassificationError}
\end{table*}

One of the errors is that the models use the plural article with singular nouns, and vice versa. For example, in the sentence \textit{Los equipo de nueva creación que quieran acceder a las ligas deberán disputar un partido amistoso} (The newly formed teams that wish to access the leagues must play a friendly match.), the determinant is in plural and the noun in singular. The language used in Llama2 is mostly English and not so much Spanish. 

A common mistake related to Easy to Read generation is to use different terms to refer to the same thing. For example, the \textit{futsal} regulations document sometimes refers to team members as "players" and other times as "athletes". 

The easy-reading adapter highlighted the importance of providing explanations or including certain terms in a glossary when working in a field with technical terminology. For example, the word \textit{acta}(score sheet) is defined in the glossary of the Gold Standard document, while the model output presents it without explanation or replaces it with the synonym \textit{lista}(roster).

The adaptations aimed at readers with difficulties have improved the Llama2 model through fine-tuning, to simplify language for people with intellectual disabilities. It has been shown that good results are achieved by using a training dataset focused on the domain of simplification. Creating a dataset that encompasses different domains adapted for easy reading could represent a valuable resource to achieve even more competitive results in other areas. This task can also be applied to English, a simpler language than Spanish, with a more complex morphology and extensive verbal inflection. The texts in Spanish are characterized by the presence of numerous subordinate clauses and extensive phrases, which can exhibit a wide variation in word order.

\subsection{Implications and Future Directions}
The conducted study opens new possibilities for future research in lexical simplification and adaptation to easy reading in Spanish, highlighting the versatility of the employed methodology, which can be applied across various fields and domains. It was observed that the Llama2 7B model provides adequate results in an initial evaluation. However, a greater number of texts adapted for easy reading could significantly enhance the obtained results. On the other hand, more advanced models such as the Llama2 70B offer even more competitive outcomes. To optimize accuracy, it is suggested to utilize the Llama2 70B model without quantization and fine-tuning using a broad set of adaptations for easy reading, potentially leading to exceptional performance with high-quality results. It's worth noting that this approach would require considerably high computational resources for implementation.

\section{Conclusion}

The research described in this article represents the first study on simplifying to Spanish Easy to Read language for people with intellectual disabilities, through the use of decoder-based LLMs. We have outlined the procedure for creating a parallel corpus of Easy to Read texts composed of a collection of 1941 sentences that is a valuable resource to train machine learning approaches to simplify content. In the near future, it would be beneficial to explore other strategies to group pieces of information (sentences, paragraphs, or even entire texts) to obtain a broader context in this type of corpus. Increasing the size of these resources is an open challenge because adapting LLM to new tasks requires larger corpora.  Some tests have been conducted with texts from domains that the model is unfamiliar with, and the results are not as good as when the model is familiar with the domain in question.

We suggest that future research consider incorporating training with a domain terminology dictionary in which researchers are currently working. This is because some documents contain complex or uncommon terminology depending on the domain being worked on. This way, when an unfamiliar term appears, it can be accurately described, facilitating reader comprehension.

The experiments presented illustrate a use case and show that the corpus has allowed the evaluation of lexical simplification approaches based on language models. The methodology used could be applied to other languages, such as English, for which linguistic models with more training data are available.

In summary, this first approach significantly improves the accessibility of documents in Spanish for people with intellectual disabilities. In addition, the development of corpora plays a crucial role in the development of simplification systems for people with reading comprehension difficulties. However, it is essential to emphasize that documents simplified through an automated system must always be reviewed and validated by a professional.

\section*{Conflict of Interest Statement}
%All financial, commercial or other relationships that might be perceived by the academic community as representing a potential conflict of interest must be disclosed. If no such relationship exists, authors will be asked to confirm the following statement: 

The authors declare that the research was conducted in the absence of any commercial or financial relationships that could be construed as a potential conflict of interest.

\section*{Author Contributions}

PM: conceptualization, investigation, methodology, writing - original draft, writing-review, and editing.\\
AR: data curation, software, validation, writing - original draft.\\
LM: conceptualization, resources, funding acquisition, writing - original draft. \\

\section*{Funding}
This work was supported by ACCESS2MEET project (PID2020-116527RB-I0) supported by MCIN AEI/10.13039/501100011033/.

\section*{Acknowledgments}
We would like to thank the adapter of the AMAS foundation for his assistance with the result evaluation.

\bibliographystyle{unsrtnat}
\bibliography{references}  %%% Uncomment this line and comment out the ``thebibliography'' section below to use the external .bib file (using bibtex) .

\end{document}